\documentclass[10pt,twocolumn,letterpaper]{article}

\usepackage{cvpr}
\usepackage{times}
\usepackage{epsfig}
\usepackage{graphicx}
\usepackage{amsmath}
\usepackage{amssymb}
\usepackage{caption}
\usepackage{subcaption}
\usepackage{booktabs}
\usepackage{relsize}

\usepackage[pagebackref=true,breaklinks=true,letterpaper=true,colorlinks,bookmarks=false]{hyperref}

\cvprfinalcopy 

\def\cvprPaperID{7638} 

\newcommand\norm[1]{\left\lVert#1\right\rVert}

\begin{document}
\title{3dDepthNet: Point Cloud Guided Depth Completion Network for Sparse Depth and Single Color Image}

\author{Rui Xiang\thanks{Paper is finished during author's internship at Trifo, Inc.}\\
Department of Mathematics\\
University of California, Irvine\\
{\tt\small xiangr1@uci.edu}
\and
Feng Zheng, Huapeng Su,  Zhe Zhang\\
Trifo, Inc.\\
{\tt\small \{feng.zheng, huapeng.su, zhe.zhang\}@trifo.com}
}

\maketitle

\begin{abstract}
   In this paper, we propose an end-to-end deep learning network named 3dDepthNet, which produces an accurate dense depth image from a single pair of sparse LiDAR depth and color image for robotics and autonomous driving tasks. Based on the dimensional nature of depth images, our network offers a novel 3D-to-2D coarse-to-fine dual densification design that is both accurate and lightweight. Depth densification is first performed in 3D space via point cloud completion, followed by a specially designed encoder-decoder structure that utilizes the projected dense depth from 3D completion and the original RGB-D images to perform 2D image completion. Experiments on the KITTI dataset show our network achieves state-of-art accuracy while being more efficient. Ablation and generalization tests prove that each module in our network has positive influences on the final results, and furthermore, our network is resilient to even sparser depth.
\end{abstract}


\section{Introduction}
\begin{figure}
\centering
\begin{subfigure}[b]{0.45\linewidth}
    \includegraphics[width=\linewidth]{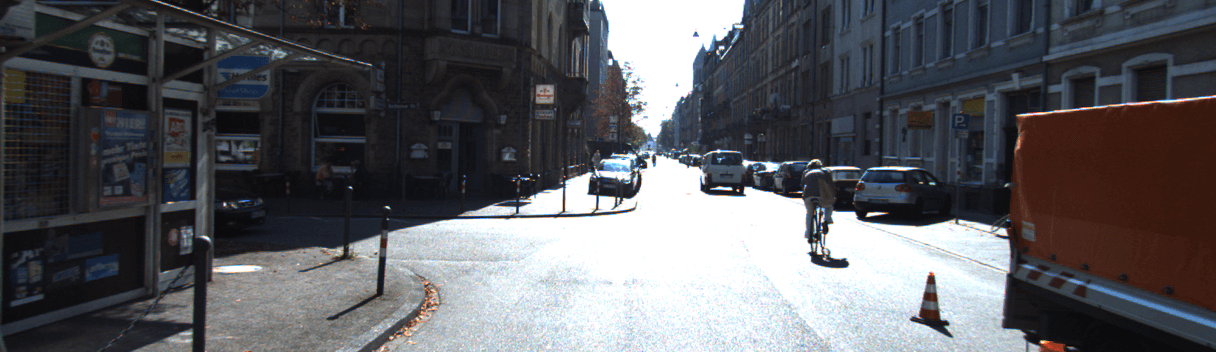}
    \caption{Color Image}
\end{subfigure}
\begin{subfigure}[b]{0.45\linewidth}
    \includegraphics[width=\linewidth]{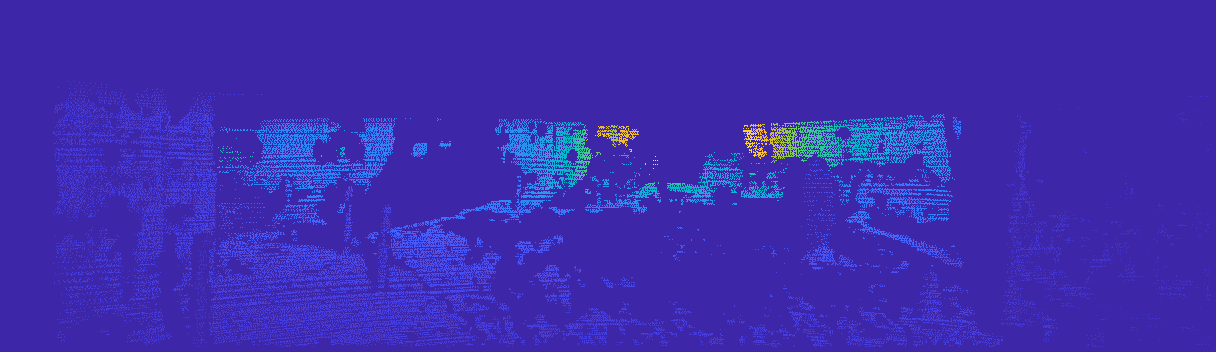}
    \caption{Groundtruth Depth}
\end{subfigure}
\begin{subfigure}[b]{\linewidth}
    \centering
\includegraphics[width=0.9\linewidth]{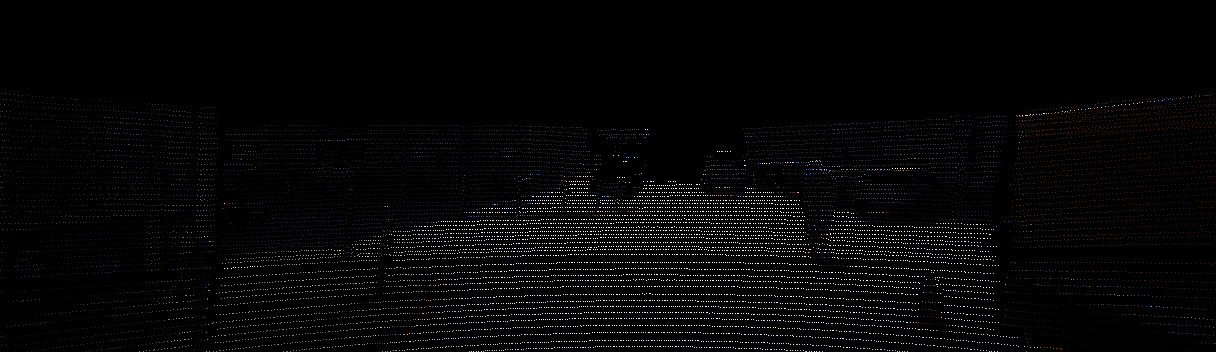}
    \vspace*{-2mm}\caption{Sparse Depth from LiDAR}
\end{subfigure}
\begin{subfigure}[b]{\linewidth}
    \centering
\includegraphics[width=0.9\linewidth]{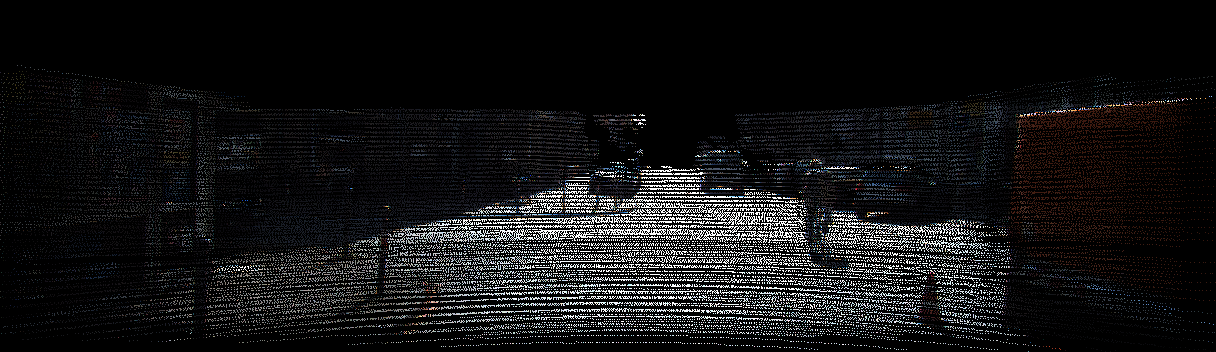}
    \vspace*{-2mm}\caption{Projected Dense Depth from LiDAR Copmletion Net}
\end{subfigure}
\begin{subfigure}[b]{\linewidth}
\centering
    \includegraphics[width=0.9\linewidth]{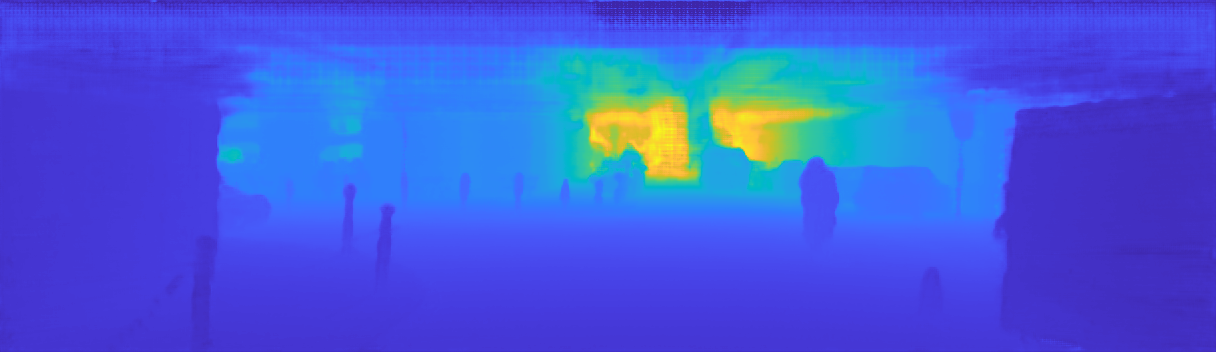}
    \vspace*{-2mm}\caption{Predicted Smooth Dense Depth}
\end{subfigure}
\caption{Our model takes sparse depth (c) as input to generate a dense point cloud by our LiDAR Completion Net (d), and then predict the smooth, dense depth (e) via our Depth Completion Net with the aid of the projected dense depth, and the original color image (a) and sparse depth (c).}
\end{figure}
Depth sensing is a fundamental task for both indoor and outdoor applications, such as home robotics and autonomous driving. Indoor depth sensors, e.g. RGB-D camera, are usually more accurate because of the limited distance of indoor scenes and good illumination conditions \cite{fanello2017ultrastereo,ryan2016hyperdepth}. For outdoor scenes, LiDAR is usually the primary depth sensor due to its high accuracy and long sensing range. However, even high-resolution LiDAR albeit being extremely expensive\footnote{Currently, the price of 16-line and 64-line Velodyne LiDARs is \$4k and \$75k, respectively \cite{ma2019self}.} still produces sparse depth output. The absence of affordable dense depth sensors leads to a great research interest to develop methods to estimate smooth and dense depth from sparse samples. 
\begin{figure*}
\begin{center}
 \includegraphics[width=\linewidth]{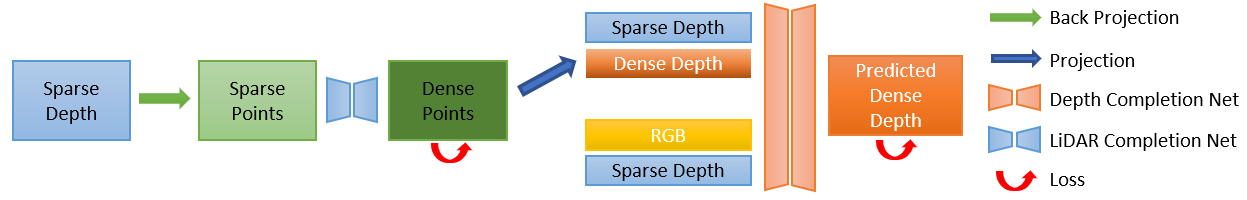}
\end{center}
   \caption{\textbf{Network architecture.} There are two important units in our network: LiDAR Completion Net and Depth Completion Net. LiDAR Completion Net performs point cloud completion in 3D space according to local geometry, while Depth Completion Net fuses projected dense depth from 3D completion and original sparse depth and color image to produce a smooth and dense depth image output.}
\label{Net}
\end{figure*}

There are two main challenges in this depth completion task. The first challenge is feature extraction from highly sparse depth measurements and in particular, how to combine features from different input modalities (i.e. depth and color). Simple concatenation may not keep consistency well since depth and color values have different units. Also, the object boundary could be fuzzy due to the occlusion problem, which is the phenomenon that the depth image may not perfectly align with the color image. Secondly, for learning-based methods, dense groundtruth depth is difficult to obtain. Compared to image classification or segmentation tasks, pixel-wise manual annotation of depth values is not reliable and extremely time-consuming.

\begin{figure*}
\begin{center}
 \includegraphics[scale=0.35]{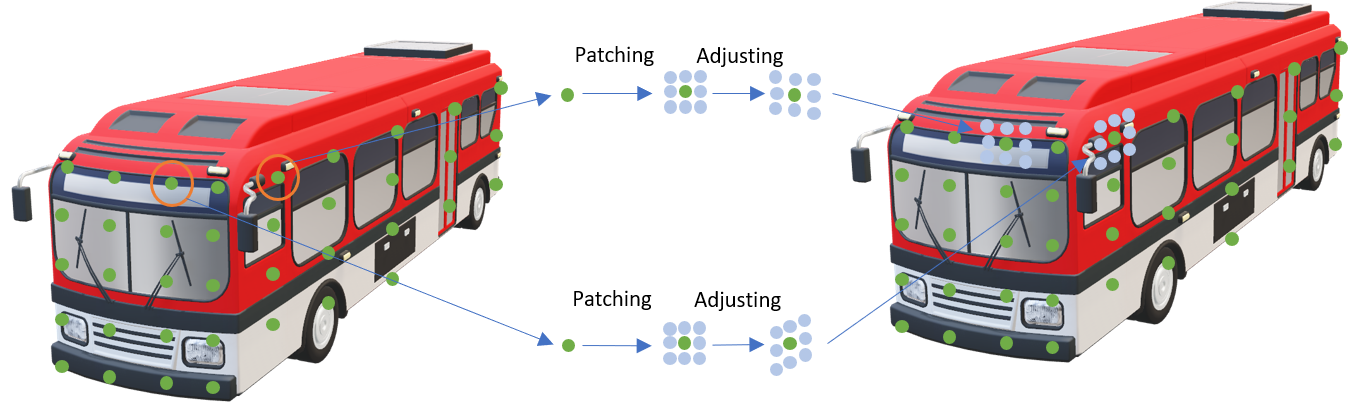}
\end{center}
   \caption{\textbf{Intuition of LiDAR Completion Net.} LiDAR Completion Net performs both patching of original points and then adjusting the patch to fit its local geometry. Green dots represent original LiDAR scan points, and blue dots represent new patch points.}
\label{LCN2}
\end{figure*}

The architecture of our proposed network is shown in Figure \ref{Net}. We summarize our main contribution as follows:   

\begin{enumerate}
  \item We propose a novel 3D-to-2D coarse-to-fine depth completion network, which performs coarse densification in 3D space via point cloud completion, followed by 2D image space fine depth completion with the aid of the generated intermediate dense 3D point cloud.
  \item Based on the property of sparse LiDAR points, we propose a LiDAR Completion Net, which performs patching, adjusting, and densification in 3D space. We illustrate the process in figure \ref{LCN2} and show that our LiDAR Completion Net performs a geometrically meaningful 3D completion.
  \item We propose a modified encoder-decoder structure, called Depth Completion Net, to fuse the output from LiDAR Completion Net and the original sparse depth and color image to generate the final dense and smooth depth image.
\end{enumerate}

\section{Related Work}
\textbf{Point Cloud Completion} Raw LiDAR scan data is represented as point cloud in 3D space. Including LiDAR data, most real-world 3D data are often incomplete, resulting in a loss in geometric information. Incompleteness mainly comes from two major limitations of modern 3D sensors, the first one is the viewing angle limitation, and the second one is the resolution limitation. Current completion methods for 3D data can be categorized into traditional methods and learning-based methods. 

For traditional methods, interpolation methods \cite{berger2014state,davis2002filling,nealen2006laplacian} are explored to fill holes or gaps in relatively small local surface. Symmetry detection methods \cite{mitra2006partial,mitra2013symmetry,pauly2008discovering} utilize symmetry property of nature objects to complete unobserved symmetrical parts. Such methods are limited by their strong assumptions at the underlying geometry of the 3D objects. Alignment-based methods \cite{han2008bottom,li2015database,karsch2012depth,kim2013learning} retrieve object or object parts from a library to match or assemble the target incomplete 3D object. Planes and quadrics are also taken into account for some alignment-based methods \cite{chauve2010robust,li2011globfit}. These methods often have to solve a large-scale optimization problem and are limited by the content of the corresponding library.

For learning-based methods, most methods use voxels as the representation in order to take advantage of the voxel convolutional neural network. However, voxel-based methods are significantly limited by its computation cost both in the discretization step and training step. There are also learning methods \cite{litany2018deformable} based on mesh deformation. \cite{yuan2018pcn} is a point cloud-based learning method dealing with specific 3D object completion. However, there are few methods designed for point cloud completion of large scale scenes, such as LiDAR scan data. The content complexity and information sparsity of LiDAR scan data make the completion task extremely difficult even for deep learning methods.

\textbf{Depth Completion} Depth completion aims to recover an accurate, smooth, and dense depth image, given a semi-dense or sparse depth image with or without the guidance of the corresponding color image. Structure light-based sensors and laser-based sensors have very different resolution and range. The sensor-dependent nature and different input modalities of depth completion make this task very challenging and application-specific. The sparser the input depth, the more difficult the problem becomes. 

For indoor applications, with structured light sensor, typically less than 20\% depth is missing. Traditional methods like depth inpainting \cite{barron2016fast, chen2017multi}, depth super-resolution \cite{mac2012patch,shabaninia2017high,yu2013shading}, and depth denoising \cite{diebel2006application} already provides good results. Deep learning methods also have been used for indoor depth completion. Zhang \textit{et al.} \cite{zhang2018deep} proposed adding surface normal information into learning network. Chen \textit{et al.} \cite{cheng2018depth} utilized an affinity matrix to guide the depth completion with a recurrent neural network.

The problem becomes more difficult when the outdoor scene is involved because the depth information is extremely sparse. The typical sensor for outdoor application, e.g., autonomous driving, is LiDAR, which provides a very limited number of line scan depth information. Even high-resolution LiDAR can only provide 4\% depth information in one image. Wavelet methods \cite{hawe2011dense, liu2015depth} are used before the emerging of deep learning. Recently, Qiu \textit{et al.} \cite{qiu2019deeplidar} proposed a two pathway deep learning network involving surface normal and attention mask to generate geometrically meaningful output. Ma \textit{et al.} \cite{ma2019self} designed an encoder-decoder network to fuse the information from RGB channel and depth channel with an extension on self-supervised learning. Uhrig \textit{et al.} \cite{uhrig2017sparsity} proposed a sparse convolution filter to address the importance of data sparsity. Chodosh \textit{et al.} \cite{chodosh2018deep} combined deep learning and dictionary learning into an integrated network to generate dense depth. Chen \textit{et al.} \cite{chen2019learning} proposed a domain-specific network that applies 2D convolution on image pixels and continuous convolution on 3D points.

Most state-of-art methods tackle the problem in 2D image space via CNN. In fact, the essence of the depth image is 3D but not 2D. Our method takes advantage of this intrinsic property of depth image by performing completion in both 3D point cloud space and 2D image space sequentially, which makes the completion more geometrically meaningful. To our best knowledge, our 3dDepthNet is the first work that integrates 3D point cloud completion into 2D depth image completion problem.

\textbf{Depth Prediction} Depth prediction tasks predict depth from a single monocular color image. With the absence of depth measurements, depth prediction is even more challenging than depth completion. Handcrafted features \cite{karsch2012depth, karsch2014depth,saxena2006learning} are used in the early study of this problem. Recent learning-based methods also integrated those features into novel loss functions to achieve a better result. For example, Zhou \textit{et al.} \cite{zhou2017unsupervised} proposed to use the photometric loss as supervision during training. Mahjourian \textit{et al.} \cite{mahjourian2018unsupervised} and Yin \textit{et al.} \cite{yin2018geonet} added 3D geometric constrains to predict depth and estimate optical flow. Qi \textit{et al.} \cite{qi2018geonet} combined estimated surface normal and depth prediction into a two pathway neural network. 

Even though most approaches achieve promising results on prediction, the scale ambiguity problem still exists because no accurate depth guidance is provided. When it comes to industrial applications, the performance of depth prediction based on a single monocular camera might not be stable.

\section{The Proposed Method}
Unlike state-of-art depth completion works which function in 2D image space via CNN \cite{qiu2019deeplidar, ma2019self, van2019sparse, xu2019depth}, we lift the problem to 3D since depth image, in essence, is three dimensional. Specifically, we propose an end-to-end deep learning network to complete the depth in 3D point cloud space and image space sequentially. The network takes sparse depth image, RGB image, and corresponding calibration information\footnote{Since the calibration information is already used when projecting raw LiDAR data to the image space, we do not consider it as extra information.} as input to generate a dense depth image.

Completion is first performed in 3D space by patching sparse 3D LiDAR points and adjusting the position of corresponding patch points (LiDAR Completion Net). After projecting the dense point cloud onto image space, we pass it through a specially designed encoder-decoder structure (Depth Completion Net) to fuse the information from the original sparse depth image, projected dense depth image, and RGB image to produce a smooth dense depth image. The whole structure is illustrated in Figure \ref{Net}.

\subsection{LiDAR Completion Net}
\label{sec:lcn}
\begin{figure}
\begin{center}
 \includegraphics[width=0.9\linewidth]{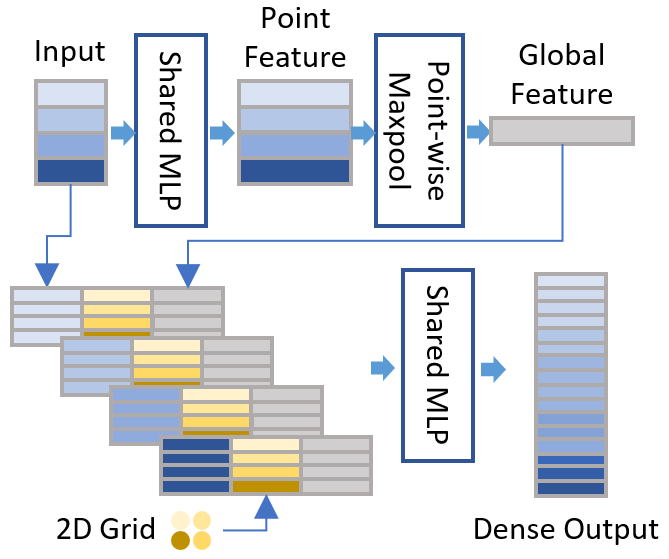}
\end{center}
   \caption{\textbf{LiDAR Completion Net.} We first extract global features by PointNet, and then concatenate the 2D grid coordinates with repeated local point coordinates and global features. Next, combined local and global features are fed into a shared-MLP to produce dense point cloud output. Each colored row in the above figure stands for one point in the corresponding feature space; the same color represents the same point.}
\label{LCN1}
\end{figure}

\begin{figure*}
\begin{center}
        \begin{subfigure}[b]{0.25\textwidth}
                \includegraphics[width=\linewidth]{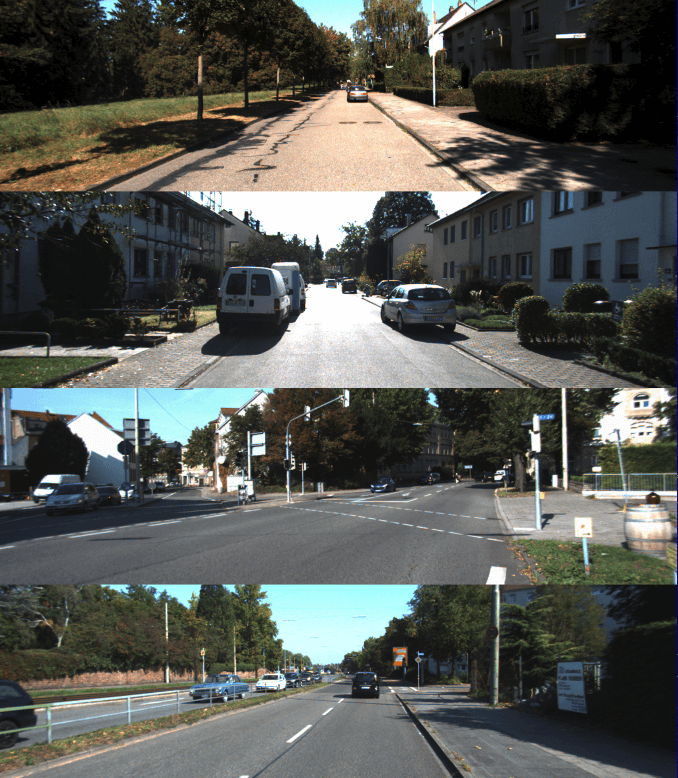}
                \caption{RGB Image}
        \end{subfigure}%
        \begin{subfigure}[b]{0.25\textwidth}
                \includegraphics[width=\linewidth]{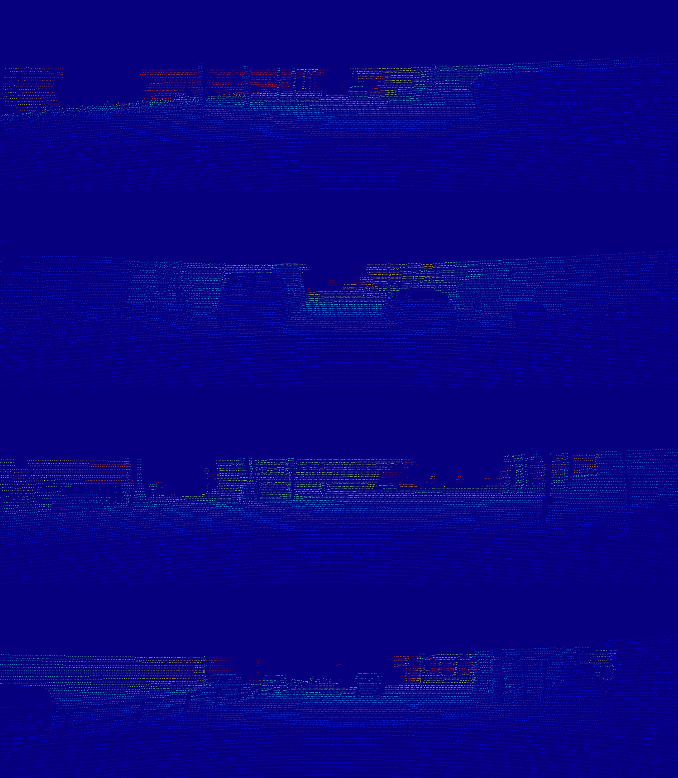}
                \caption{Sparse Depth}
        \end{subfigure}%
        \begin{subfigure}[b]{0.25\textwidth}
                \includegraphics[width=\linewidth]{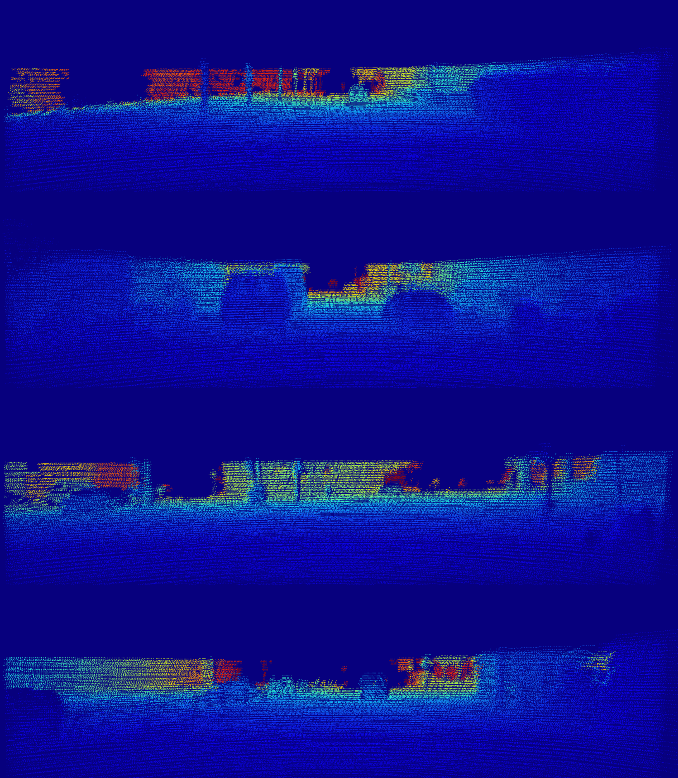}
                \caption{Projected Dense Depth}
        \end{subfigure}%
        \begin{subfigure}[b]{0.25\textwidth}
                \includegraphics[width=\linewidth]{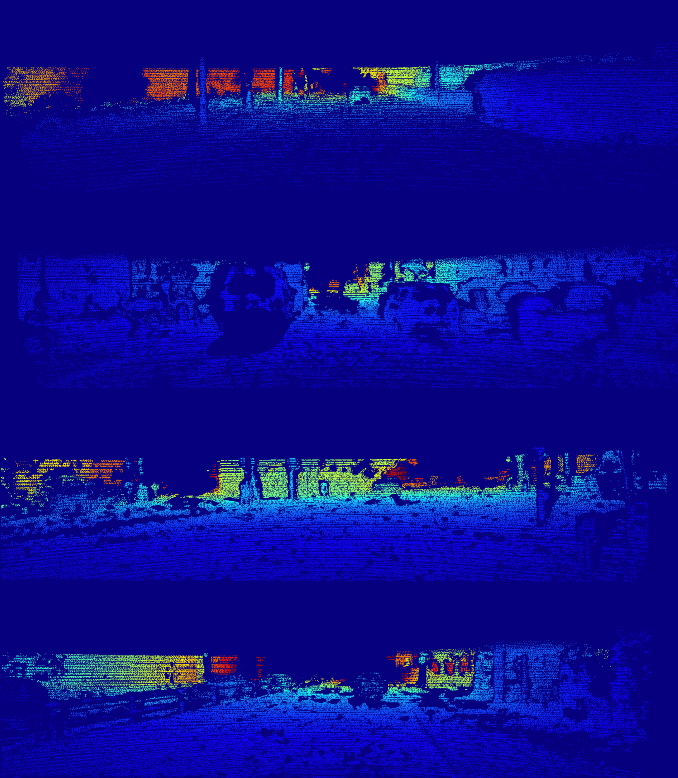}
                \caption{Ground Truth}
        \end{subfigure}%

\end{center}
        \caption{\textbf{Visualization of the LiDAR Completion Net.} To visualize the result of LiDAR Completion Net, we project the dense point cloud to the image space and compare it with groundtruth depth image. The third column is the projection result of our dense point cloud generated by the LiDAR Completion Net.}
\label{lcn_res}
\end{figure*}

Completion in 3D space is a long-studied subject. Learning-based methods \cite{yang2018foldingnet, yuan2018pcn, sung2015data} mainly focus on single object point cloud completion instead of the entire scene in our scenario. In our work, the sparse depth information serves as landmark points in 3D space, and we aim to complete the missing neighboring depth information around those anchor points. Inspired by the Point Completion Net\cite{yuan2018pcn} (PCN) which is designed for single object point cloud completion, we propose a LiDAR Completion Net by adjusting the structure of PCN, making it more suitable for sparse LiDAR points completion task. The shared-MLP forming coarse level points in PCN is removed, where coarse level points are directly replaced by sparse input points. Moreover, the second half of the encoder in PCN is removed. These changes make our LiDAR Completion Net more efficient and effective in completing sparse LiDAR points, as shown in Figure~\ref{lcn_res}.

As illustrated in Figure \ref{LCN1} and \ref{LCN2}, sparse LiDAR points are first fed into a PointNet \cite{qi2017pointnet} to extract global feature vector. Meanwhile, raw LiDAR points are also served as coarse level landmarks. Global feature vector, the local landmark, and patch points around the corresponding landmark are then combined into a multi-level shared-MLP to generated dense point clouds. 

By directly using raw LiDAR points as coarse level landmarks, the number of parameters used in our LiDAR Completion Net is much less than PCN since most parameters are used for generating coarse level landmarks in original PCN, PointNet encoder and last shared-MLP only have very fewer parameters. In our experiments, LiDAR Completion Net only has around 0.2 million parameters while original PCN has 6.85 million parameters.

\subsection{Depth Completion Net}
\begin{figure}
\begin{center}
 \includegraphics[width=0.9\linewidth]{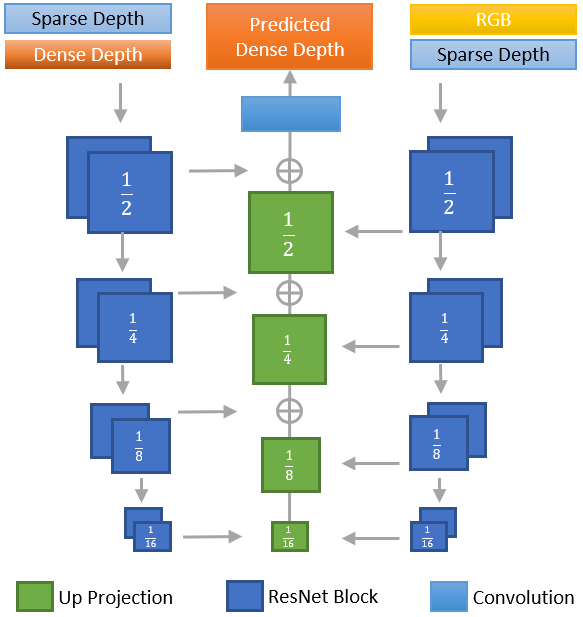}
\end{center}
   \caption{\textbf{Depth Completion Net.} Similar to \cite{qiu2019deeplidar}, we use late fusion strategy, where different encoder features are only combined in the decoder stage. To emphasize the importance of depth information, we sum the dual-depth channel and concatenate RGB-D channel in the decoder.}
\label{DCN1}
\end{figure}

The Depth Completion Net is always the core part of methods solving depth completion problem in image space. We propose a network architecture that fuses the information of projected dense depth image from our LiDAR Completion Net, original sparse depth image, and RGB image, as shown in Figure \ref{DCN1}.

Inspired by \cite{qiu2019deeplidar}, our Depth Completion Net consists of two pathways; one is a dual-depth pathway that integrates completed depth from LiDAR Completion Net and sparse depth, the other one is a RGB-D pathway which takes original RGB image and sparse depth as input. The left pathway is the main pathway, while the right pathway provides guidance during the up-projection stage to generate smoother results. Encoders in both pathways consist of ResNet blocks to obtain a 1/16 downsized feature. The decoder then gradually integrates features from both pathways and increases feature resolution back to the original resolution. As suggested in \cite{qiu2019deeplidar}, we concatenate features from the right pathway but sum the features from the left path. The decoder is encouraged to keep consistency and learn more from the completed depth and the original sparse depth when features from the left pathway are added to it \cite{chen2017multi}. Different from \cite{qiu2019deeplidar} which has three encoder-decoder structures, we only need one encoder decoder, which results in a significant decrease in the total number of parameters. Comparing with one encoder-decoder network such as \cite{ma2019self} with ResNet34 \cite{he2016deep} as building blocks, we only use ResNet18 which is half of the parameter numbers used in ResNet34. Therefore, our Depth Completion Net is more lightweight and computationally efficient.

\subsection{Loss Function} \label{train strategy}

Our loss function is defined as:
\begin{equation}
    L = \{L_{pt}, L_{im}\}
\end{equation}
where $L_{im}$ is the standard masked MSE loss between generated dense depth image and ground truth depth image, and $L_{pt}$ is the Chamfer Distance \cite{fan2017point} between the dense point cloud generated from LiDAR Completion Net and ground truth point cloud.  

The Chamfer Distance between two point clouds $P_1$ and $P_2$ is defined as:
\begin{equation}
\begin{split}
    CD(P_1, P_2) = & \frac{1}{|P_1|} \sum_{x\in P_1} \min_{y\in P_2} \norm{x-y}_2 \\
    &+ \frac{1}{|P_2|} \sum_{y\in P_2} \min_{x\in P_1} \norm{x-y}_2
\end{split}
\end{equation}
It computes the average closest point distance between the $P_1$ and $P_2$. Since the parameter number and loss magnitude are very different in $L_{pt}$ and $L_{im}$, we train our network in two steps. We first freeze parameters in the Depth Completion Net and train parameters in the LiDAR Completion Net for 1 epoch, and then we freeze parameters in the LiDAR completion Net and train parameters in Depth Completion Net for 11 epochs. We use Adam as our optimizer with an initial learning rate 0.001, $\beta_1=0.9$, $\beta_2=0.999$ and drop learning rate to half for every 2000 steps during training LiDAR Completion Net, for every 5 epoch during training Depth Completion Net.

\section{Experiments}
In this section, we first describe the training dataset and device. Next, we compare our method against existing state-of-art methods and conduct an ablation study of our method. Finally, we show the potential generalization ability of our model.

\subsection{Training Data}
Our model is trained on KITTI depth completion benchmark \cite{Geiger2013IJRR}. Original KITTI data set provides RGB images, sparse depth images, raw LiDAR scans, and calibration files. We generate point cloud files from raw sparse depth images with corresponding calibration files. We use 1 TITAN RTX GPU for training and 1 RTX 2080Ti GPU for testing. Since we have limited computation resources, the training is on one camera's data from scratch and fine-tune on both cameras' data.
\subsection{Comparison with State-of-art Methods}
\textbf{Metrics.} Following the KITTI benchmark, we use 4 standard metrics to measure the quality of our method: root mean square error in $mm$ (RMSE), mean absolute error in $mm$ (MAE), root mean squared error of the inverse depth in $1/km$ (iRMSE) and mean absolute error of the inverse depth in $1/km$ (iMAE). iRMSE and iMAE compute the mean error of inverse depth and concentrate more on close objects while RMSE and MAE directly measure the accuracy of all estimated depth and focus more on further objects. RMSE is used as the dominant metric in the KITTI benchmark leader board. 

\textbf{Evaluation on KITTI Test set.}
KITTI test set contains 1,000 RGB images and corresponding sparse depth. The groundtruth is not provided; all test results will be evaluated on the KITTI test server to prevent overfitting. 

\begin{figure*}
\begin{center}
        \begin{subfigure}[b]{0.25\textwidth}
                \includegraphics[width=\linewidth]{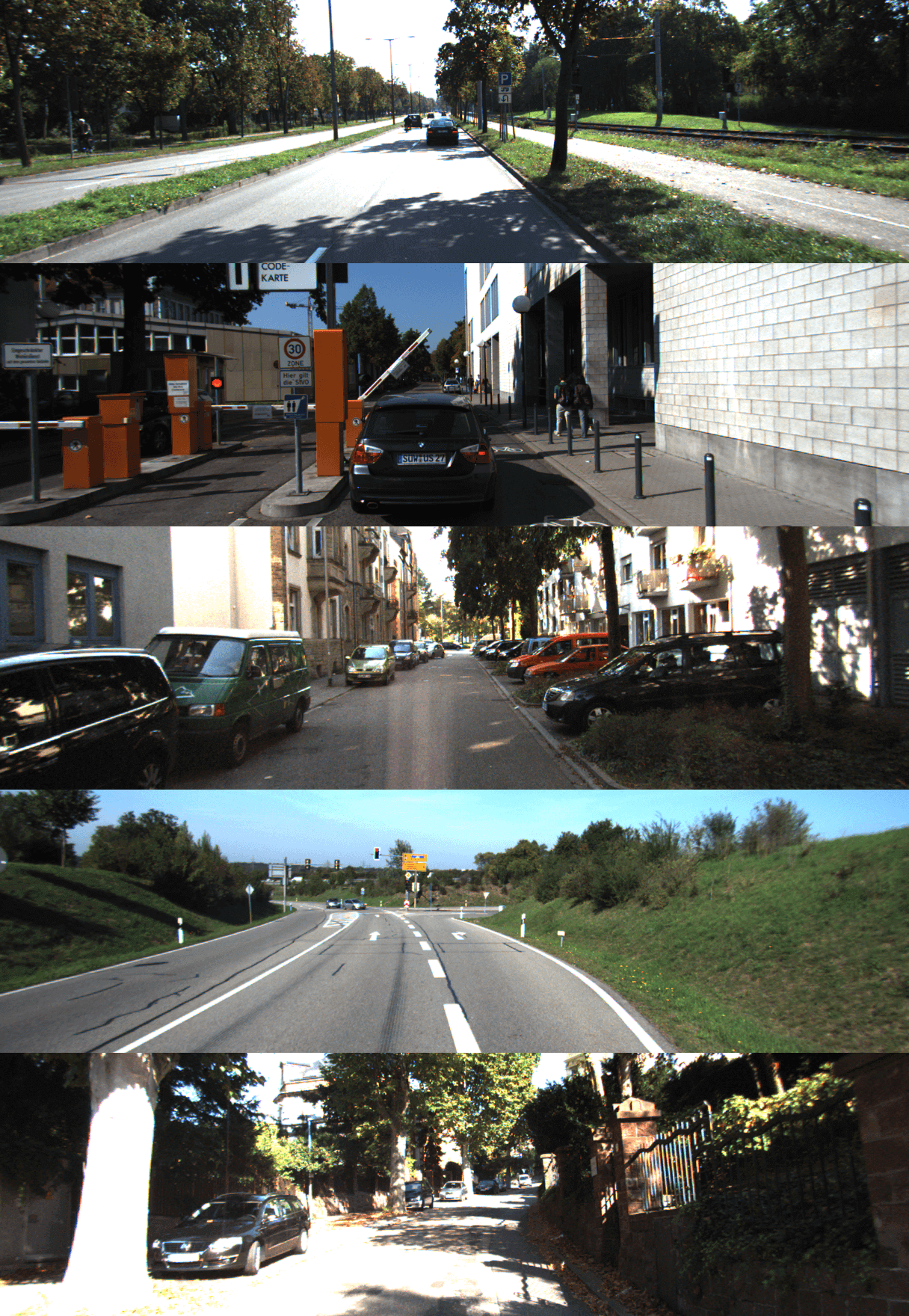}
                \caption{RGB Image}
        \end{subfigure}%
        \begin{subfigure}[b]{0.25\textwidth}
                \includegraphics[width=\linewidth]{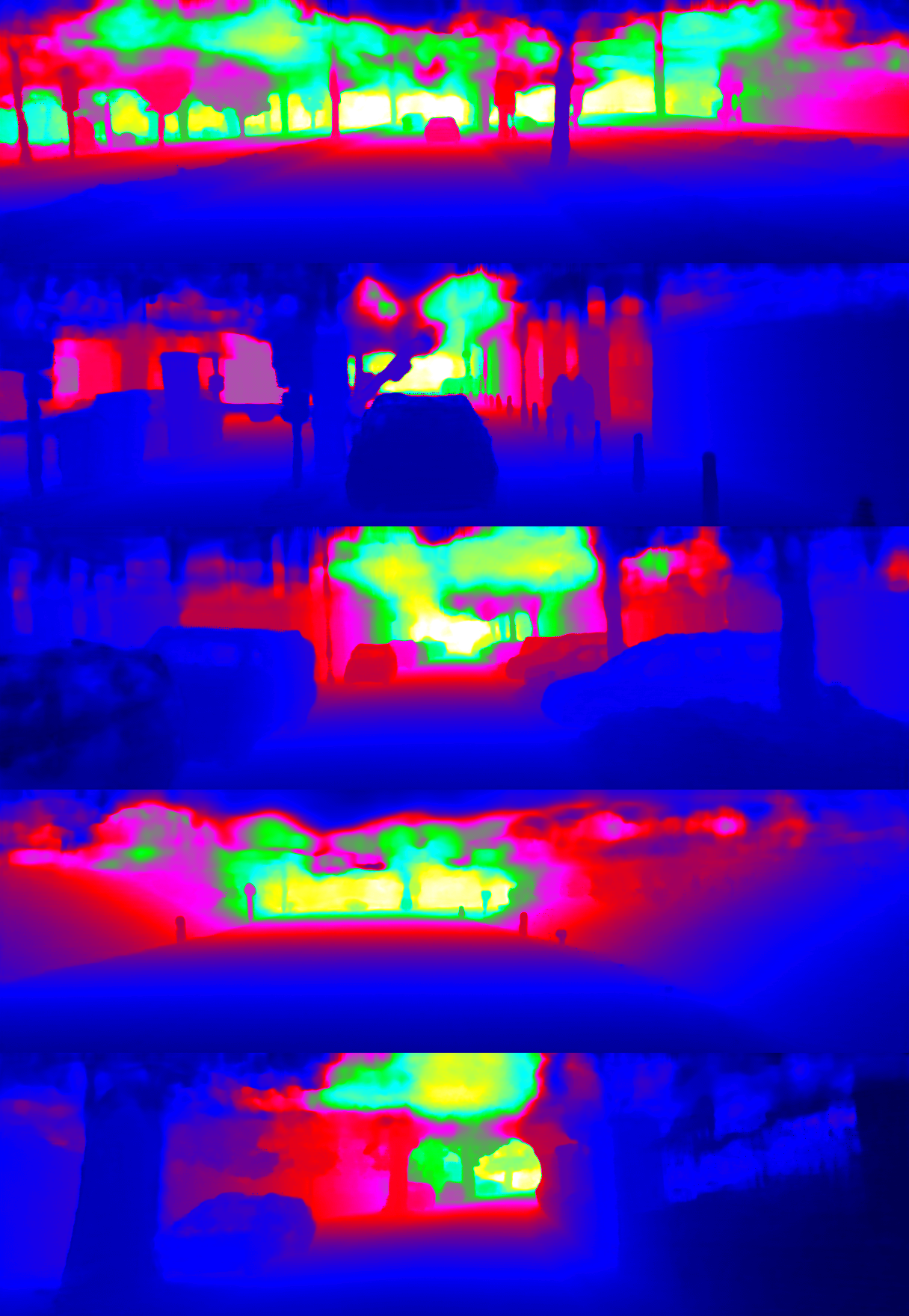}
                \caption{PwP \cite{xu2019depth}}
        \end{subfigure}%
        \begin{subfigure}[b]{0.25\textwidth}
                \includegraphics[width=\linewidth]{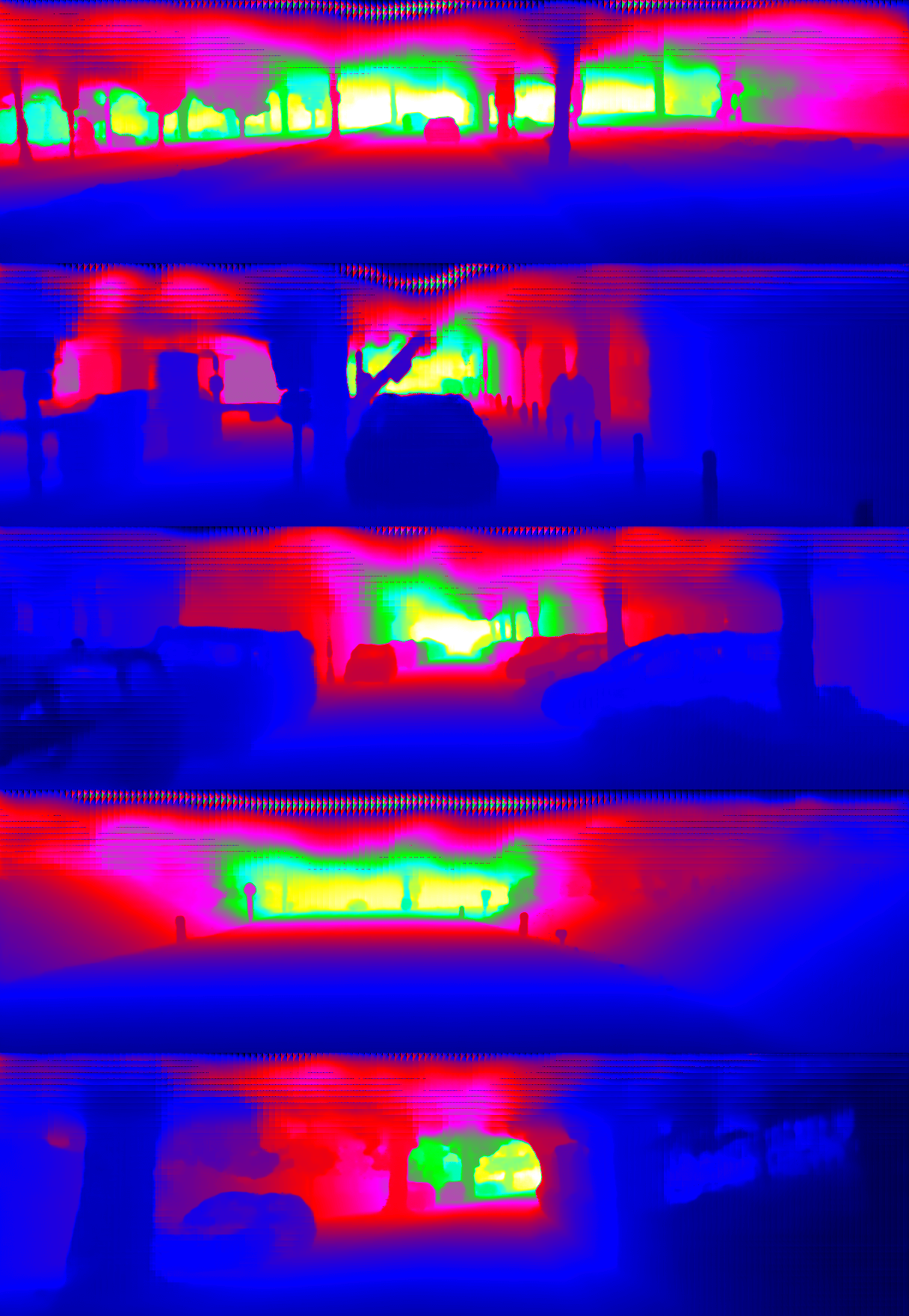}
                \caption{DeepLiDAR \cite{qiu2019deeplidar}}
        \end{subfigure}%
        \begin{subfigure}[b]{0.25\textwidth}
                \includegraphics[width=\linewidth]{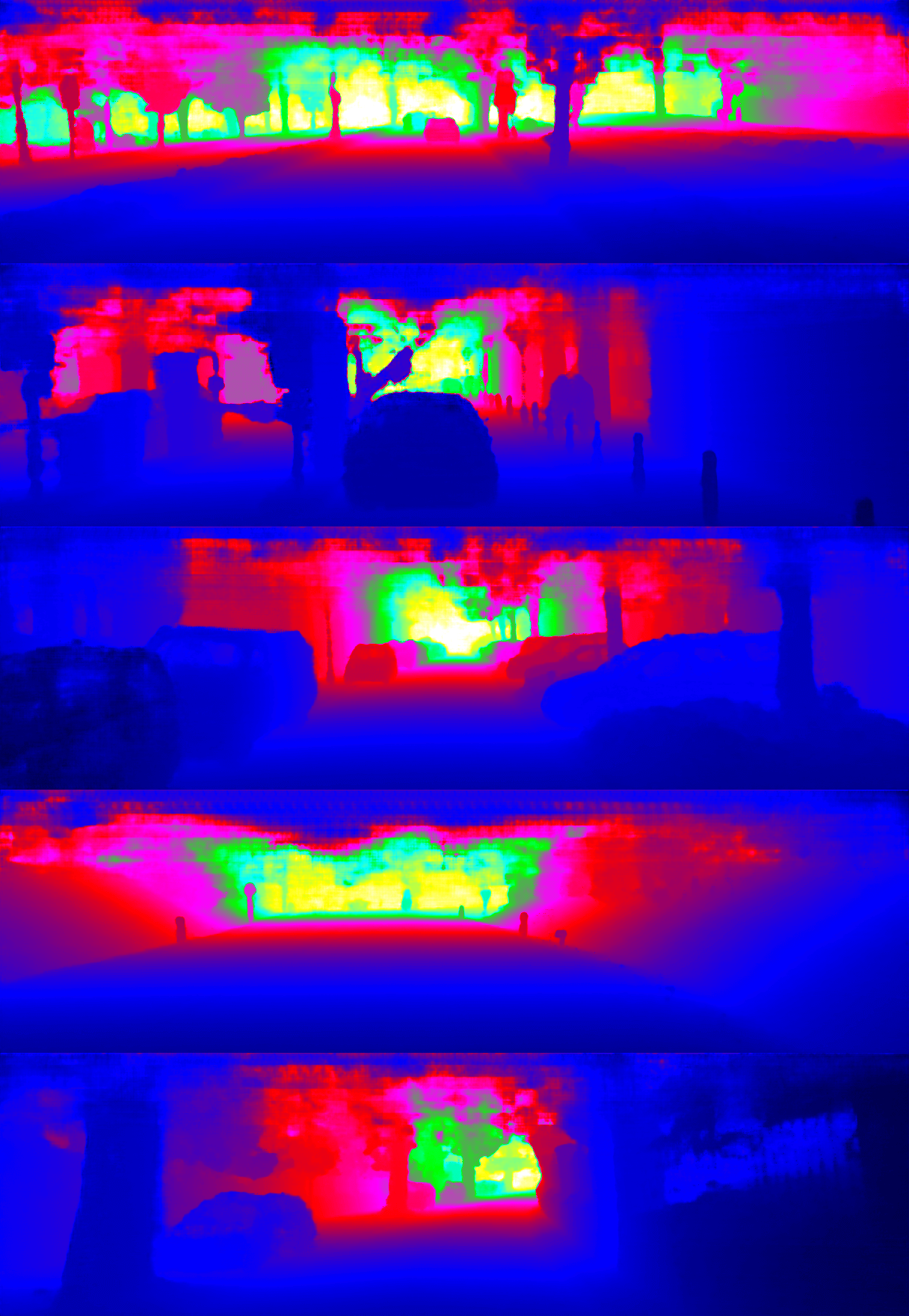}
                \caption{Ours}
        \end{subfigure}%
\end{center}
\caption{\textbf{Visualization Comparison on KITTI Test Set} We list the visualization results of PwP \cite{xu2019depth}, DeepLiDAR \cite{qiu2019deeplidar} and our method of several test images. Our method is more accurate on close objects as indicated by lower iRMSE and iMAE. For areas without groudtruth depth guidance, such as upper part in color image, our network produces more meaningful results than PwP and has less aliasing than DeepLiDAR.} 
\label{test_res}
\end{figure*}

Quantitative results of our method and other state-of-art methods are listed in Table \ref{test_test}. Qualitative results are shown in Figure \ref{test_res}. Our method produces more stable and accurate details on close objects than most of the other methods as indicated by lower iRMSE and iMAE. 

Moreover, since our LiDAR Completion Net and Depth Completion Net are more lightweight compared to other state-of-art methods, our network is more efficient. We list the device and runtime comparison with several state-of-art methods in Table \ref{rt}.

\begin{table}[]
\begin{center}
\scalebox{0.9}{
\begin{tabular}{ccccc}
\hline
                                     & RMSE                        & MAE                         & iRMSE                     & iMAE \\ \hline
\multicolumn{1}{l|}{HMS-Net \cite{huang2018hms}}         & \multicolumn{1}{c|}{841.78} & \multicolumn{1}{c|}{253.47} & \multicolumn{1}{c|}{2.73} & 1.13 \\
\multicolumn{1}{l|}{MSFF-Net \cite{wang2018multi}}         & \multicolumn{1}{c|}{836.69} & \multicolumn{1}{c|}{241.54} & \multicolumn{1}{c|}{2.63} & 1.07 \\
\multicolumn{1}{l|}{NConv-CNN \cite{eldesokey2018propagating}}       & \multicolumn{1}{c|}{829.98} & \multicolumn{1}{c|}{233.26} & \multicolumn{1}{c|}{2.60} & 1.03 \\
\multicolumn{1}{l|}{Sparse-to-Dense \cite{ma2019self}} & \multicolumn{1}{c|}{814.73} & \multicolumn{1}{c|}{249.95} & \multicolumn{1}{c|}{2.80} & 1.21 \\
\multicolumn{1}{l|}{PwP \cite{xu2019depth}}       & \multicolumn{1}{c|}{777.05} & \multicolumn{1}{c|}{235.17} & \multicolumn{1}{c|}{2.42} & 1.13 \\
\multicolumn{1}{l|}{DeepLiDAR \cite{qiu2019deeplidar}}       & \multicolumn{1}{c|}{\textbf{758.38}} & \multicolumn{1}{c|}{226.50} & \multicolumn{1}{c|}{2.56} & 1.15 \\\hline
Ours                           & 798.44                       & \textbf{226.27}                       & \textbf{2.36}                      & \textbf{1.02} \\ \hline
\end{tabular}}
\caption{\textbf{Comparison on KITTI Test Set.} We selected several published state-of-art methods listed on the KITTI leaderboard. The evaluation is done on KITTI testing server. Our method outperforms these methods on iRMSE, MAE, iMAE and achieves comparable RMSE}
\label{test_test}
\end{center}
\end{table}

\begin{table}[]
\begin{center}
\scalebox{0.85}{
\begin{tabular}{lccc}
\hline
\multicolumn{1}{c}{}                            & Device                          & Runtime                   & \begin{tabular}[c]{@{}c@{}}Converted \\ Run Time\end{tabular} \\ \hline
\multicolumn{1}{l|}{DeepLiDAR \cite{qiu2019deeplidar}} & \multicolumn{1}{c|}{GTX 1080Ti} & \multicolumn{1}{c|}{0.07s} & 0.07s              \\
\multicolumn{1}{l|}{Sparse-to-Dense \cite{ma2019self}} & \multicolumn{1}{c|}{Tesla V100} & \multicolumn{1}{c|}{0.08s} & 0.16s              \\
\multicolumn{1}{l|}{PwP \cite{xu2019depth}}            & \multicolumn{1}{c|}{Tesla V100} & \multicolumn{1}{c|}{0.1s}  & 0.19s              \\ \hline
\multicolumn{1}{c}{Ours}                        & RTX 2080Ti                      & 0.03s                      & 0.05s              \\ \hline
\end{tabular}
}
\end{center}
\caption{\textbf{Runtime Comparison on the KITTI validation set.} We list inference speeds and corresponding devices of several state-of-art methods and ours. According to \cite{lambda}, RTX 2080Ti has 1.62 times average speedup over GTX 1080Ti on FP16 computation, and Tesla V100 has 1.98 times average speedup. Hence, our method is the most efficient one with the consideration of different devices.}
\label{rt}
\end{table}

\textbf{Evaluation on KITTI Validation set.}
We further compare our method with some other methods which are not on the KITTI benchmark on the KITTI validation set. KITTI validation set also contains 1,000 RGB images and corresponding sparse depth image with the ground truth dense depth image provided. Compared methods include bilateral filter using color (Bilateral) \cite{tomasi1998bilateral}, fast bilateral (Fast) \cite{durand2002fast}, optimization using total variance (TGV) \cite{ferstl2013image}, and deep depth completion for indoor scene \cite{zhang2018deep}. The first three methods are non-learning based methods, which do not perform very well because of the huge complexity of the depth completion model. Indoor learning-based methods also do not perform very well because of the scene change. The quantitative result is in Table \ref{vali_test}.

\begin{table}[]
\begin{center}
\scalebox{0.9}{
\begin{tabular}{ccccc}
\hline
                               & RMSE                         & MAE                          & iRMSE                      & iMAE  \\ \hline
\multicolumn{1}{l|}{Bilateral\cite{tomasi1998bilateral}} & \multicolumn{1}{c|}{2989.02} & \multicolumn{1}{c|}{1200.56} & \multicolumn{1}{c|}{9.67}  & 5.08  \\
\multicolumn{1}{l|}{Fast\cite{durand2002fast}}      & \multicolumn{1}{c|}{3548.87} & \multicolumn{1}{c|}{1767.80} & \multicolumn{1}{c|}{26.48} & 9.13  \\
\multicolumn{1}{l|}{TGV\cite{ferstl2013image}}       & \multicolumn{1}{c|}{2761.29} & \multicolumn{1}{c|}{1068.69} & \multicolumn{1}{c|}{15.02} & 6.28  \\
\multicolumn{1}{l|}{Zhang $et\ al.$\cite{zhang2018deep}}     & \multicolumn{1}{c|}{1312.10} & \multicolumn{1}{c|}{356.60}  & \multicolumn{1}{c|}{4.29}  & 1.41  \\ \hline
Ours                           & \textbf{693.23}                       & \textbf{208.96}                       & \textbf{2.37}                      & \textbf{0.98} \\ \hline
\end{tabular}}
\caption{\textbf{Comparison on KITTI Validation Set.} Our method outperforms three non-learning based methods and one indoor learning based method in all metrics.}
\label{vali_test}
\end{center}

\end{table}

\subsection{Ablation Study}
\begin{figure*}
\begin{center}
 
         \begin{subfigure}[b]{0.25\textwidth}
                \includegraphics[width=\linewidth]{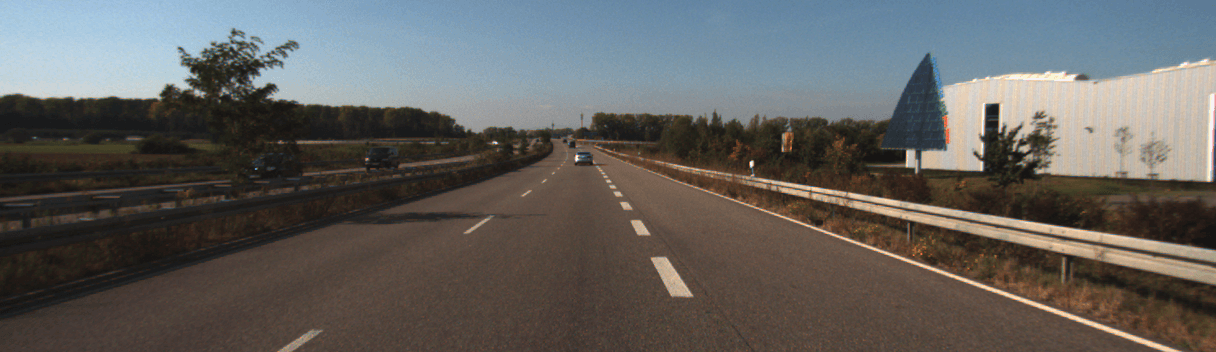}
                \caption{RGB Image}
        \end{subfigure}%
        \begin{subfigure}[b]{0.25\textwidth}
                \includegraphics[width=\linewidth]{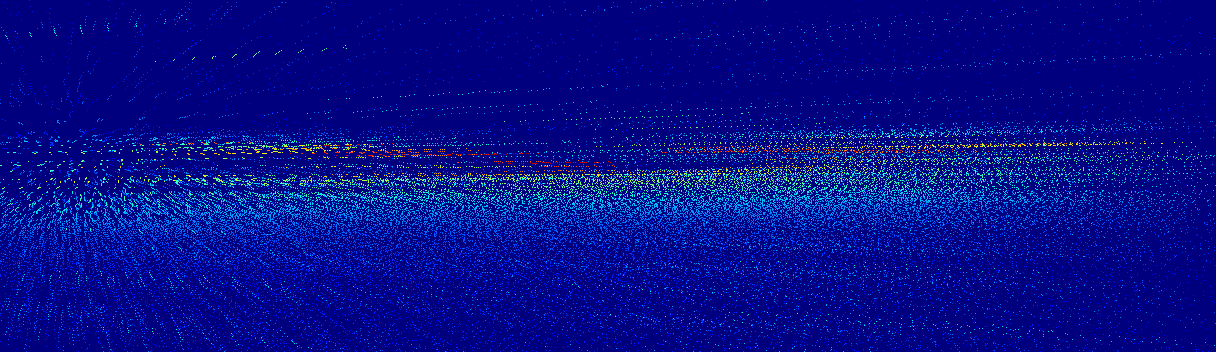}
                \caption{PCN}
        \end{subfigure}%
        \begin{subfigure}[b]{0.25\textwidth}
                \includegraphics[width=\linewidth]{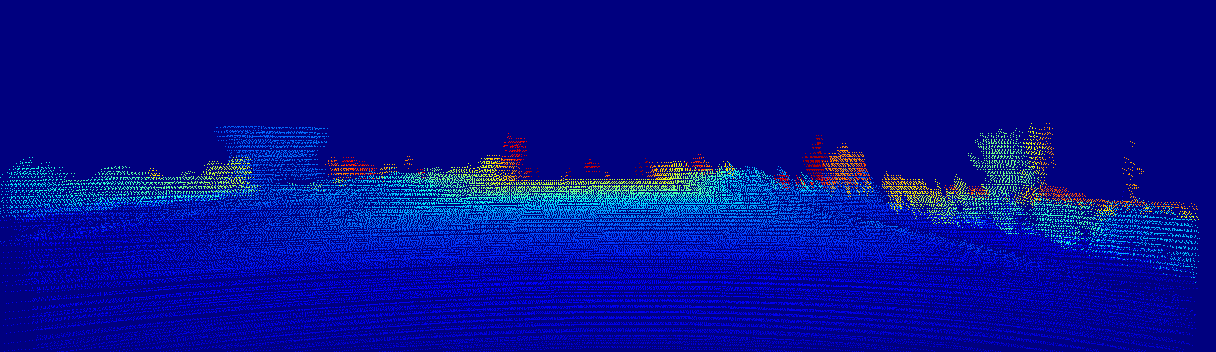}
                \caption{LiDAR Completion Net}
        \end{subfigure}%
        \begin{subfigure}[b]{0.25\textwidth}
                \includegraphics[width=\linewidth]{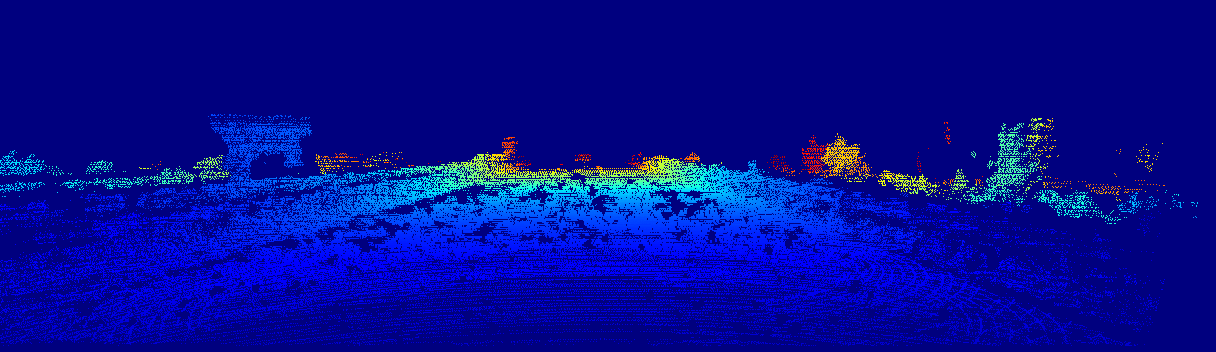}
                \caption{Ground Truth}
        \end{subfigure}%
\end{center}
   \caption{\textbf{Comparison of PCN and LiDAR Completion Net.} We visualize the results of the retrained PCN and LiDAR Completion Net for 3D scene point cloud completion. LiDAR Completion Net produces more meaningful projected dense depth.}
\label{pcn_c}
\end{figure*}

To better understand the impact of our network modules, we conduct a systematical ablation study by numerically presenting the influence of disabling specific components in our network. The quantitative result is listed in Table \ref{ab_test}, and the qualitative result of LiDAR Completion Net is shown in Figure \ref{pcn_c}.

\textbf{Effectiveness of LiDAR Completion Net.} The dense depth image generated from the LiDAR Completion Net is an essential part of our network. To verify the effectiveness of it, we replace the dense depth in the dual depth channel by the original sparse depth. The Depth Completion Net will be the only component left, so it is equivalent to study the ability of Depth Completion Net. We train the modified network with the same training strategy described in Section \ref{train strategy}. All the metrics drop significantly. This indicates that the dense depth image generated from LiDAR Completion Net provides additional important information which sparse depth image and RGB image can not provide. This extra information helps improve the final dense depth in all metrics. 

Furthermore, in order to show our modification to PCN as described in Section~\ref{sec:lcn} is effective, we compare the performance of the PCN \cite{yuan2018pcn} and our LiDAR Completion Net on scene point cloud completion. For simplicity, we only train both of them to perform 3D point cloud completion with randomly selected 5,000 pairs of data from KITTI training set for 10 epochs. The Depth Completion Net is removed and only the Chamfer Distance loss is back-propagated. Completed point clouds are projected to image space for visualization. As shown in figure \ref{pcn_c}, our LiDAR Completion Net is much more suitable for completing sparse point cloud of an entire scene.  

\textbf{Effectiveness of Depth Completion Net.} We verify the positive effects of our Depth Completion Net by replacing it with the image encoder-decoder structure describe in \cite{ma2019self}. The significant difference of \cite{ma2019self} with our Depth Completion Net is that it only has one encoder pathway, and all the skip connection is implemented by concatenation. We change the sparse depth input into a two-channel depth by concatenating the dense depth generated from LiDAR Completion Net. The training strategy is the same as described in Section \ref{train strategy}.

\begin{table*}[]
\begin{center}
\begin{tabular}{lcccccc}
\hline
                             & LiDAR Completion Net     & Depth Completion Net     & RMSE                        & MAE                         & iRMSE                     & iMAE  \\ \hline
\multicolumn{1}{l|}{Model 1} & \multicolumn{1}{c|}{No}  & \multicolumn{1}{c|}{Yes} & \multicolumn{1}{c|}{767.68} & \multicolumn{1}{c|}{249.07} & \multicolumn{1}{c|}{3.67} & 1.14  \\
\multicolumn{1}{l|}{Model 2} & \multicolumn{1}{c|}{Yes} & \multicolumn{1}{c|}{No}  & \multicolumn{1}{c|}{742.28} & \multicolumn{1}{c|}{221.29} & \multicolumn{1}{c|}{2.52} & 1.10  \\ \hline
\multicolumn{1}{l|}{Full}    & \multicolumn{1}{c|}{Yes} & \multicolumn{1}{c|}{Yes} & \multicolumn{1}{c|}{693.23} & \multicolumn{1}{c|}{208.96} & \multicolumn{1}{c|}{2.37} & 0.98 \\ \hline
\end{tabular}
\end{center}
\caption{\textbf{Ablation study on KITTI validation set.}}
\label{ab_test}
\end{table*}

\subsection{Generalization}
\begin{figure}
\centering
    \includegraphics[width=\linewidth]{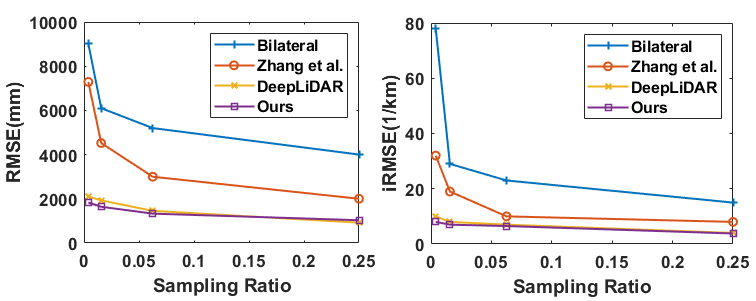}
\caption{\textbf{Generalization test on different sparsity input.} Due to our novel design of the 3D-to-2D coarse-to-fine densification pipeline, in particular our LiDAR Completion Network, our model generalizes well on sub-sampled sparse depth.}
\label{gtest}
\end{figure}

We further test the generalization ability of our method by providing sparser input. In the KITTI data set, the original LiDAR depth provides around 4\% depth information which is close to 17,000 pixels with depth in a 1216 by 352 image. To test the generalization ability of our network, We sub-sample the depth information in sparse depth data from LiDAR to 1/4, 1/16, 1/64, and 1/256. The quantitative result of RMSE and iRMSE comparing with several other methods is illustrated in Figure \ref{gtest}. The performance of our network drops as the sparsity increases as expected. Our network outperforms other methods when the sampling ratio becomes lower. The patching process in LiDAR Completion Net helps densify the sparse LiDAR data, and this phenomenon amplifies when sampling density drops. This explains why our network is more resilient to even sparser depth.

\section{Discussion}

\textbf{3D point cloud completion.} The LiDAR Completion Net is an important part of our network architecture. How to perform a better 3D completion task for large scale point cloud data is essential to generate a dense and smooth depth image. Our LiDAR Completion Net performs \textit{patching} and \textit{adjusting} functions according to global feature and the coordinate of each sparse point. To further improve the quality of LiDAR Completion Net, local geometry information and semantic information can be taken into consideration. Object segmentation information in 3D space can also help perform an object-specific point cloud completion. Specifically, points describing people or cars should be treated differently in LiDAR Completion Net.

\begin{figure}
\begin{center}
 \includegraphics[width=0.9\linewidth]{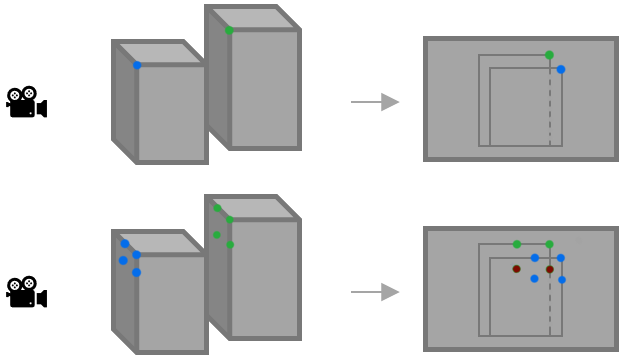}
\end{center}
   \caption{\textbf{Penetration Problem.} The first row demonstrates the process from LiDAR scan points (first column) to depth image (second column), which has no penetration problem since the LiDAR senor can not penetrate through solids. The second row is the dense point cloud generated by LiDAR Completion Net. After projection, the two red dots will exist in the depth image since blue points can not block them. However, for LiDAR scan depth image, red dots are not allowed to show up in the smaller rectangular region according to the perspective principle.}
\label{pena}
\end{figure}

\textbf{Penetration Problem.}
Once the completion is finished in 3D space, the dense point cloud will be projected into image space to generate a dense depth image. However, since objects described by point cloud are not solid, points describing faraway objects may penetrate through close object point cloud onto the projection plane, as illustrated in Figure \ref{pena}. The original sparse depth image does not have this problem because the LiDAR scan can not penetrate through solid objects. In our network, the Depth Completion Net learns to solve the penetration problem. However, we feel like there is no geometric explanation and theoretical guarantee on how good this problem is solved by learning. We aim to solve it more elegantly in future work. 

\textbf{Combination of 2D and 3D completion.}
Most previous methods to depth completion employ 2D image encoder-decoder structures that take color and sparse depth image as input. In our case, with the generated intermediate dense point cloud from Lidar Completion Net, how to design an encoder-decoder structure that efficiently utilizes RGB image, sparse depth, and intermediate dense point cloud, becomes essential. Our Depth Completion Net with dual depth channel plus RGB-D channel design is modified from 2D image encoder-decoder structure, which only takes image input. We will explore possibilities of designing a new encoder-decoder structure that directly uses information from 3D point cloud and 2D images.

\section{Conclusion}

We propose a new approach to the depth completion problem by introducing 3D completion into the learning network. Our novel deep learning network offers a 3D-to-2D coarse-to-fine dual densification design. By taking advantage of the dimensional nature of depth image, we propose a novel LiDAR Completion Net to do completion in 3D space and pass it to a specially designed Depth Completion Net which integrates projected dense depth, sparse depth and RGB image to produce smooth dense depth image. Experiments show our network is efficient and achieves state-of-art accuracy. Ablation and generalization tests prove that each module in our network has positive influences on the final results, and furthermore, our network is resilient to even sparser depth.

{\small
\bibliographystyle{ieee_fullname}
\bibliography{egbib}
}

\end{document}